\documentclass[10pt,twocolumn,letterpaper]{article}

\usepackage{iccv}
\usepackage{times}
\usepackage{epsfig}
\usepackage{graphicx}
\usepackage{amsmath}
\usepackage{amssymb}

\graphicspath{{Visualization/}}

\usepackage[T1]{fontenc}

\usepackage{amsmath,amssymb,amsfonts} 
\usepackage{subcaption}
\usepackage{fixltx2e}
\usepackage{multirow}
\usepackage{booktabs}
\usepackage{amsmath}

\usepackage{algorithm,algpseudocode}

\algrenewcommand\algorithmicforall{\textbf{foreach}}

\algdef{SE}[SUBALG]{SubAlgorithm}{EndSubAlgorithm}{\algorithmicsubalgorithm}{\algorithmicend\ \algorithmicsubalgorithm}%

\algtext*{SubAlgorithm}
\algtext*{EndSubAlgorithm}% If you want to avoid seeing "end sub-algorithm"

\algdef{SE}[SUBALG]{Indent}{EndIndent}{}{\algorithmicend\ }%
\algtext*{Indent}
\algtext*{EndIndent}
\usepackage[pagebackref=true,breaklinks=true,letterpaper=true,colorlinks,bookmarks=false]{hyperref}

\iccvfinalcopy % *** Uncomment this line for the final submission

\def\iccvPaperID{3006} % *** Enter the ICCV Paper ID here

\ificcvfinal\fi
\begin{document}

%%%%%%%%% TITLE
\title{ID-aware Quality for Set-based Person Re-identification}

\author{
	Xinshao Wang\textsuperscript{1,2}, Elyor Kodirov\textsuperscript{2}, Yang Hua\textsuperscript{1,2}, Neil M. Robertson\textsuperscript{1,2} \\
	\textsuperscript{1} School of Electronics, Electrical Engineering and Computer Science, Queen's University Belfast, UK \\
	\textsuperscript{2} Anyvision Research Team, UK\\
	\{xwang39, y.hua, n.robertson\}@qub.ac.uk, \{elyor\}@anyvision.co
	\vspace{-0.5cm}
}
\date{}

\maketitle
%\thispagestyle{empty}

%%%%%%%%% ABSTRACT
\begin{abstract}
	Set-based person re-identification (SReID) is a matching problem that aims to verify whether two sets are of the same identity (ID). Existing SReID models typically generate a feature representation per image and aggregate them to represent the set as a single embedding. However, they can easily be perturbed by noises -- perceptually/semantically low quality images -- which are inevitable due to imperfect tracking/detection systems, or overfit to trivial images. In this work, we present a novel and simple solution to this problem based on ID-aware quality that measures the perceptual and semantic quality of images guided by their ID information. Specifically, we propose an ID-aware Embedding that consists of two key components: (1) Feature learning attention that aims to learn robust image embeddings by focusing on `medium' hard images. This way it can prevent overfitting to trivial images, and alleviate the influence of outliers. (2) Feature fusion attention is to fuse image embeddings in the set to obtain the set-level embedding. It ignores noisy information and pays more attention to discriminative images to aggregate more discriminative information. Experimental results on four datasets show that our method outperforms state-of-the-art approaches despite the simplicity of our approach.
	
	%	xtensive experiments on four public set-based person re-identification datasets show the advantages of IDAE compared with previous sophisticated attentive spatiotemporal models. 
\end{abstract}

%%%%%%%%% BODY TEXT

%\vspace{-2pt}
\section{Introduction}
\label{sec:introduction}
Set-based person re-identification (SReID) \cite{zheng2016mars,liu2017qan,song2017region,li2018diversity} is a matching problem that targets identifying the same person across multiple non-overlapping cameras. Each person is represented by a set consisting of multiple images. %, i.e., matching a probe person against a bunch of gallery people captured by different cameras to produce a ranked list. 
There has been an increasing attention recently because of its critical applications in video surveillance, e.g., airport and shopping mall.

% v 2.0 
%The objective of set-based verification is to determine whether two image sets are of the same object. %For example, in the case of set-based face verification (face image sets) \cite{liu2017qan,xie18a}, the target is to verify whether two image sets show the same person or not. 
%Depending on application, it could be person, face or even car. In this paper, we work on set-based person re-identification %(entire person image sets) 
%\cite{zheng2016mars,liu2017qan,song2017region,li2018diversity}, whose objective is the same as set-based face verification.  
%
%%% background / impact
%Specifically, person re-identification (ReID) targets identifying the same person across multiple non-overlapping cameras, i.e., matching a probe person against a bunch of gallery people captured by different cameras to produce a ranked list. 
%Person ReID has been attracting increasing attention because of its critical applications in video surveillance of public scenarios such as airports and shopping malls. 

\begin{figure}[!t]
	\centering
	\begin{subfigure}[b]{0.47\textwidth}
		\includegraphics[width=\textwidth]{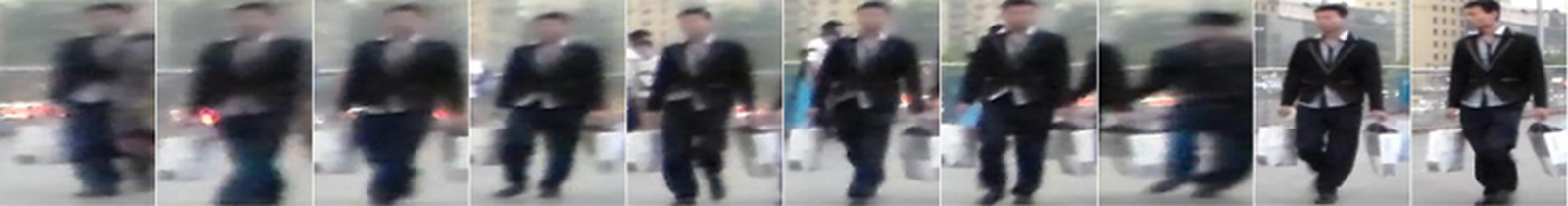}
		\vspace{-16pt}
		\caption{with/without blur}
		\vspace{3pt}
		\label{fig:quality_blur}
	\end{subfigure}
	\begin{subfigure}[b]{0.47\textwidth}
		\includegraphics[width=\textwidth]{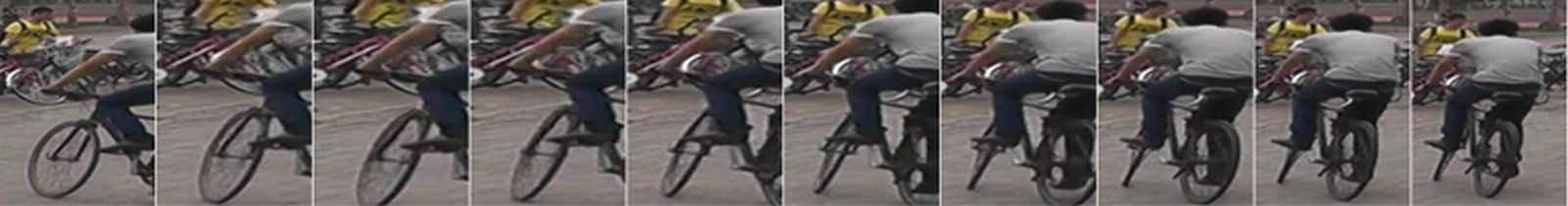}
		\vspace{-16pt}
		\caption{complete/incomplete body}
		\vspace{3pt}
		\label{fig:quality_missing_part}
	\end{subfigure}
	\hfill
	~ %add desired spacing between images, e. g. ~, \quad, \qquad, \hfill etc. 
	\begin{subfigure}[b]{0.47\textwidth}
		\includegraphics[width=\textwidth]{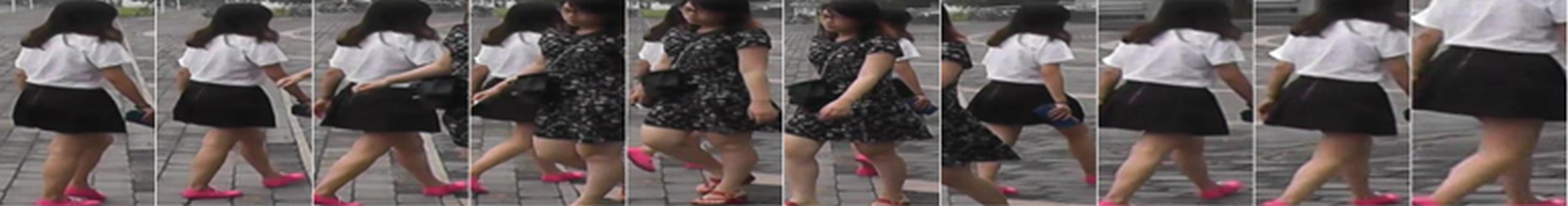}
		\vspace{-16pt}
		\caption{with/without occlusion}
		\vspace{3pt}
		\label{fig:quality_occlusion}
	\end{subfigure}
	\begin{subfigure}[b]{0.47\textwidth}
		\includegraphics[width=\textwidth]{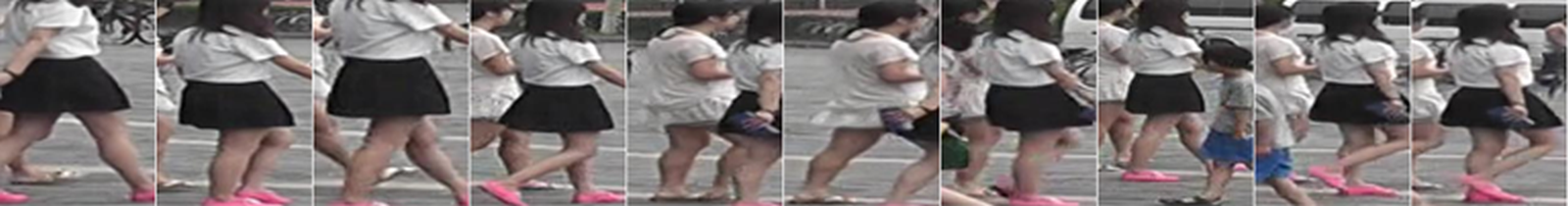}
		\vspace{-16pt}
		\caption{single/multiple people in one image}
		\vspace{3pt}
		\label{fig:quality_multiple persons}
	\end{subfigure}
	%\vspace{-16pt}
	\caption{
		Problems in set-based person re-identification. Four sets of examples are shown (a-d); each set corresponds to a particular problem. They are grouped into two:  perceptual quality and semantic quality problems. (a) belongs to the first one while the remaining ones belong to another. Note that some images in (b), (c) and (d) are less semantically related to its set ID due to incomplete body, heavy occlusion by a different person, and muptiple people in an image.  		 
		%The illustration of image's ID-aware quality. 
		%
		%Overall, ID-aware quality measures the quality of an image based on its semantic relation to its set ID. 
		%
		%It covers perceptual quality problems such as blur in set (a), as well as semantic quality problems such as incomplete body, occlusion by a different person, and multiple people in one image in set (b), (c), and (d) respectively.   
		%
		%Obviously, images with one of these problems are less semantically related to its set ID. 		
	}
	\label{fig:quality_analysis}
	%\vspace{-0.5cm}
\end{figure}
%%%%%%%%%%%%%%%%%%%%%%%%%%%%%%%%%%%%%%%%%%

%%%%%%%%%%%%%%%%%%%%%%%%%%%%%%%%%%%%%%%%%%%%%%%%%%%%

%%%%%%%%%%%%%%
%zhou2017see,xu2017jointly,
% limitation of existing methods 
Although many SReID approaches exist \cite{liu2015spatio,yan2016person,you2016top,zhu2016video,mclaughlin2017video,liu2017video,chung2017two}   
, they mainly follow two steps: feature representation and feature fusion. At high level, the feature representation is learned with a deep convolutional neural network (CNN), and then they are aggregated by simple average fusion of image feature representations in the set. However, since the average fusion treats all images equally in the set, it ignores the fact that some images are more informative than the others in the set. To this end, some SReID methods \cite{zhou2017see,xu2017jointly,liu2017qan,song2017region,li2018diversity} applied attentive aggregation in which they modify CNN such that it can generate the attention score for each image to estimate the quality, e.g., image quality estimation network \cite{liu2017qan}. Their main purpose is to identify low quality/non-discriminative image embeddings (features) and ignore them from the set at the fusing stage. Nevertheless, with regard to the learning strategy, the attention is learned implicitly without any extra supervision, instead only guided by the standard loss function. As a result, in this paper we argue that these quality-based methods address \textit{perceptual quality problems}, e.g., blurry images (Fig.~\ref{fig:quality_blur}), but they cannot solve \textit{semantic quality problems} which could be `incomplete body', `total occlusion', and `mutiple people in one image' as shown in Fig.~\ref{fig:quality_missing_part}, Fig.~\ref{fig:quality_occlusion}, and Fig.~\ref{fig:quality_multiple persons}, respectively. The semantic quality problems are inevitable in real-life since we do not have perfect detection and tracking  systems yet. 

%Most current SReID approaches  
%\cite{liu2015spatio,yan2016person,you2016top,zhu2016video,mclaughlin2017video,liu2017video,chung2017two}
%obtain set representations by average fusion of image features in the set without differentiating the images. In other words, these methods treat all images in the set equally. 
%Recently, some set-based ReID methods
%\cite{zhou2017see,xu2017jointly,liu2017qan,song2017region,li2018diversity}
%apply attentive aggregation of image features and achieve great success. 
%Generally, these methods have a net branch for generating the attention score of each image, i.e., image quality estimation branch. 
%Their main purpose is to identity and ignore non-discriminative image features when fusing image features in the set. 
%However, the automatic  attention learning is learned implicitly without extra reference/supervision and only guided by the final loss function, e.g., contrastive loss. 
%As a result, these attention-based approaches are proposed for addressing perceptual quality problems as they claimed and cannot solve other severe semantic quality problems as shown in Fig.~\ref{fig:quality_occlusion},~\ref{fig:quality_missing_part},~\ref{fig:quality_multiple persons}.
Estimating the perceptual and semantic quality  of images in the set is non-trivial. Based on our observations from Fig.~\ref{fig:quality_analysis}, we find that both kinds of quality could be estimated by how much the image is related to its set ID. To this end, we define  \textit{ID-aware quality} that measures the perceptual quality and semantic quality of images guided by their ID information. To be more specific, let us have a look at two examples in Fig.~\ref{fig:quality_analysis}. The leftmost image which is blurry in Fig. \ref{fig:quality_blur} is of less perceptual quality thus less related to its set ID -- low ID quality. The middle image in Fig.~\ref{fig:quality_occlusion} is of high perceptual quality because it is  clear (no blur). However, its semantic quality is low due to the occlusion by a different person -- low ID quality again. Thus we argue that ID-aware quality covers more diverse quality issues. Moreover, the ID label for the middle image in Fig.~\ref{fig:quality_occlusion} is wrong with respect to set ID. The existing methods \cite{song2017region,li2018diversity} which rely on ID classification loss suffer from such kind of corrupted labels.  

To realise ID-aware quality, we propose to use the classification confidence score. Our intuition is that in general high quality images (i.e., trivial images) are easily classified and get high confidence scores (e.g., 0.99) while low quality images (e.g., outliers) are less related to its set ID and get low confidence scores (e.g., 0.01). Similarly, `medium' quality images get  medium confidence scores. Furthermore, we find that there are two attractive properties of utilising classification confidence to estimate ID-aware quality. The first is that ID-aware quality can be estimated by readily available ID information -- no need for extra annotations. The next is that it is easy to obtain as ID classification loss is commonly used in most existing SReID methods.

Based on ID-aware quality, we formulate an ID-aware Embedding (IDE) to learn a robust embedding function.  IDE consists of two main components: (1) Feature learning attention (FLA) that targets learning robust image embeddings by focusing on `medium' quality images (i.e., medium hard images). This is because high quality images (i.e., trivial images) contribute little to the loss and very small gradients while low quality images (very hard images or outliers) contribute large but misleading gradients. In this case,  it can prevent overfitting to trivial images, and alleviate the influence of outliers. (2) Feature fusion attention (FFA) is to fuse image embeddings in the set to obtain the set-level embedding. It ignores noisy information and pays more attention to discriminative images to aggregate more discriminative information. IDE is an end-to-end framework built on CNN optimised by ID classification loss and set-based verification loss jointly. It is worth mentioning that it is simple as we learn global representations of images and aggregate them into a singe embedding without applying sophisticated attentive spatiotemporal features \cite{li2018diversity}. \\

%To make use of ID-aware quality for learning robust image and set embeddings, we propose ID-aware Attentive Embedding (IDAE) combining Feature Learning Attention (FLA) and Feature Fusion Attention (FFA). 
%The illustration of FLA and FFA is shown in Fig.~\ref{fig:motivation}.
%FLA targets learning robust image representations by focusing on medium quality images (i.e., medium hard images). This is because high quality images (i.e., trivial images) contribute little to the loss and very small gradients, and low quality images (very hard images or outliers) contribute large but misleading gradients.
%FFA is proposed for fusing image representations in the set and accumulates discriminative information by emphasizing on high-quality images. % Their feature vectors are discriminative and close to their ID context vectors, thus owning high classification confidences.   

% contribution 
\noindent
\textbf{Contributions}%\footnote{Our source code will be publicly available for the ease of reproduction.}
: (1) A novel concept named ID-aware quality based on the classification confidence is proposed for the set-based person re-identification. It can estimate not only the perceptual quality, but also the semantic quality of images with respect to the set ID. (2) To show the applicability of ID-aware quality, we formulate ID-aware Embedding (IDE) which learns robust set-level embeddings that can be used for set-based person re-identifcation.  Our model can be trained in an end-to-end manner. (3) Extensive experiments are carried out on four benchmarks for set-based person re-identification: MARS \cite{zheng2016mars}, iLIDS-VID \cite{wang2014person}, PRID-2011 \cite{hirzer2011person}, and LPW \cite{song2017region}. Our method achieves new state-of-the-art performance on all datasets. In addition, cross-dataset experiments are conducted, which also achieves state-of-art performance.

%\begin{itemize}
%	\item A novel concept named ID-aware quality based on the classification confidence is proposed for the set-based person re-identification. It can estimate not only the perceptual quality, but also the semantic quality of images with respect to the set ID.
%	\item To show the applicability of ID-aware quality, we formulate ID-aware Embedding (IDE) which learns robust set-level embeddings that can be used for set-based person re-identifcation.  Our model can be trained in an end-to-end manner.
%	\item Extensive experiments are carried out on four benchmarks for set-based person re-identification: MARS \cite{zheng2016mars}, iLIDS-VID \cite{wang2014person}, PRID-2011 \cite{hirzer2011person}, and LPW \cite{song2017region}. Our method achieves new state-of-the-art performance on all datasets. In addition, cross-dataset experiments are conducted, which also achieves state-of-art performance. \\ 
%\end{itemize}

%%%%%%%%%%%%%%
\begin{figure*}[t!]
	\centering
	\includegraphics[width=\linewidth]{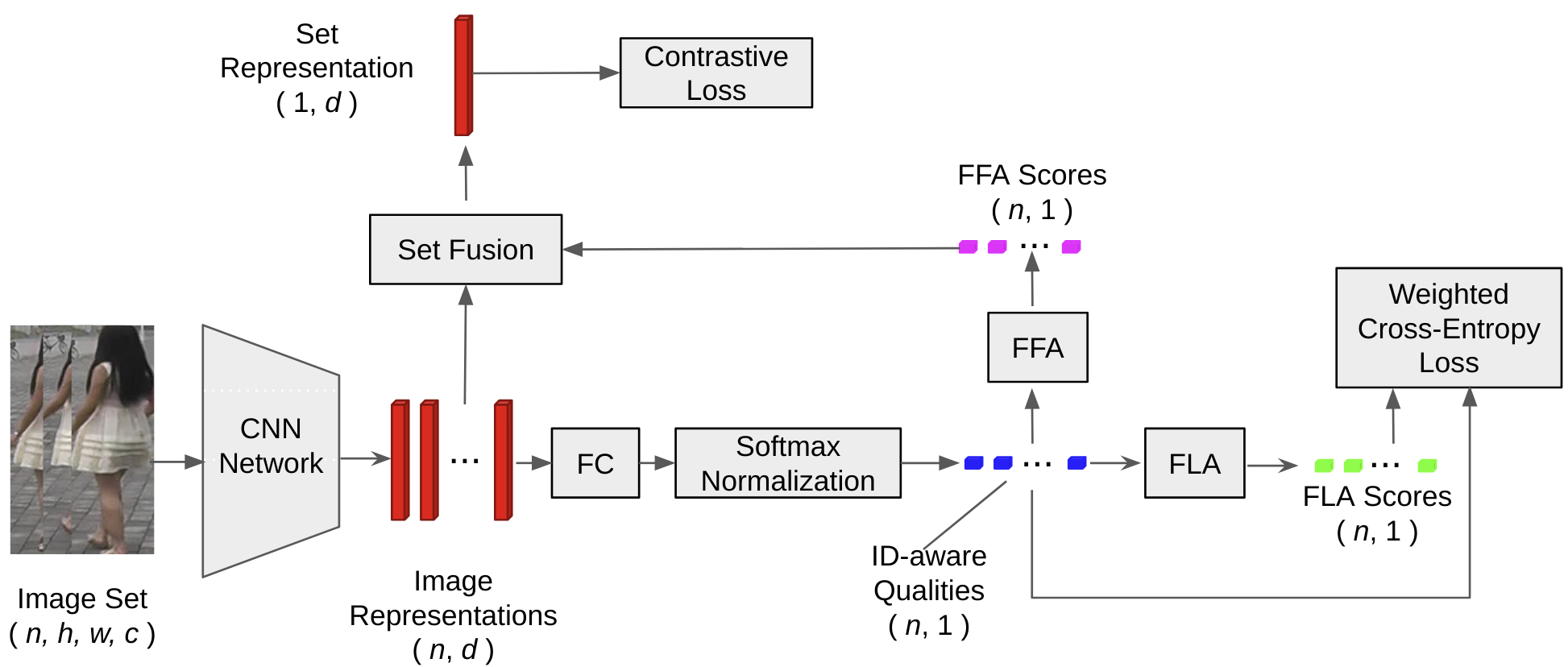}
	\caption{Overall architecture of ID-aware Embedding (IDE). An image set goes through CNN network and outputs the feature representation for each image in the set. It is followed by a fully connected layer (FC) and softmax normalization to generate ID-aware qualities. Then, these qualities are used in two ways: (1) Feature learning attention (FLA) component utilises the qualities such that medium hard images get more weights. This ends up with weighted cross-entropy loss for image-based classification. (2) Feature fusion attention (FFA) component transforms the qualites so that the ones with noise are given less weights to aggregate more discriminative information. This is supervised by contrastive loss for set-based verification. IDE is trained end-to-end and optimised by two losses jointly. 
		%FC represents fully connected layer. Both FLA and FFA generate attention scores from ID-aware qualities (classification confidences).  Weighted cross-entropy loss is applied for image-based ID classification while contrastive loss is used for set-based verification. IDAE is trained end-to-end and optimized by two losses jointly.
	}
	\label{fig:overall_architecture}
	%\vspace{-14pt}
\end{figure*}
%%%%%%%%%%%%%%textbftextbf

%%%%%%%%%%%%%%%%%%%%%%%%%%%%%%%%%%%%%%%%%%%%%%%%%%%%%%%%%%%%%
%\vspace{-2pt}
\section{Related Work}
\label{related_work}
%In this section we review the related work about person ReID. 
%Set-based person ReID is also known as video-based person ReID when temporal information or optical flow is applied \cite{mclaughlin2016recurrent}.

%\vspace{-2pt}
\noindent
\textbf{Learning Image Representations} Common approach to learning image representation is based on CNN with a certain kind of objective function such as identification loss and verification loss. This is mainly motivated by the image-based person re-identification methods \cite{xiao2016learning,zhong2017re,cheng2016person,zheng2016person}, where each ID has a single image instead of multiple images. Some methods learn a global feature representation at an image level \cite{liu2017qan,chen2018video} while others learn local image features to obtain more fine-grained representation \cite{song2017region,li2018diversity}. In this work, we follow the same strategy. However, the difference is that our method focuses on  `medium' quality images, while discard trivial (easy) images and outliers (very hard). \\

%Recently, ReID methods based on deep learning
%have become dominant. 

%%%%%%%%%%%%%%%%%%%%%%%%%%%%%%%%%%%%%%%%%%%%
%\vspace{-2pt}
\noindent
\textbf{Learning Set Representations} We review several approaches of learning set representations for SReID. We divide them into two groups in terms of whether they consider the quality of the images. 

\textit{Non-quality-based methods}. The methods belonging to this group do not consider quality, rather they focus more on temporal information \cite{liu2018spatial,mclaughlin2017video,liu2015spatio,you2016top,zhu2016video,yan2016person,mclaughlin2016recurrent,liu2017video,chung2017two} or designing different network architechtures based on 3D convolutional nets \cite{li2018multi}. %A lot of recent work \cite{liu2015spatio,you2016top,zhu2016video,yan2016person,mclaughlin2016recurrent,liu2017video,chung2017two}
%on SReID obtains the video representation by aggregating image representations \emph{without differentiating the images} in each video. 
For example, long short term memory (LSTM) \cite{hochreiter1997long} is applied in \cite{yan2016person} for aggregating image representations  and the output at the last time stamp is taken as the final representation of each video/set; Recurrent neural network (RNN) is applied in 
\cite{mclaughlin2017video,liu2017video,chung2017two}
for accumulating the temporal information in each sequence and spatial features of all images are fused by average pooling. %Note that set-based person re-identification is also called video-based person re-identification. we would like to clarify that in our work we do not care the order of images (frames) in the videos thus the name set. 

\textit{Quality-based methods.} The methods belonging to this group consider the quality of individual images in the set when aggregating image-level feature embeddings. The quality is formulated in the form of attention paradigm. That is, if the attention score is high for a particular image, then it is assumed that it is of high quality. 
The methods in \cite{wang2014person,huang2018video} select the most discriminative set (video) fragments to learn set representations, which can be regarded as fragment-level attention. Recently, image-level attention attracts a great deal of interest and has been widely studied \cite{zhou2017see,xu2017jointly,liu2017qan,song2017region,chen2018video,li2018diversity}.
In particular, in attentive spatial-temporal pooling networks \cite{xu2017jointly} and co-attentive embedding \cite{chen2018video}, an attention mechanism is designed such that  the computation of set representations in the gallery depends on the probe data. %They are less flexible and more computationally expensive as the representations of sets in gallery need to be computed again when a different probe comes. 
A similar approach is taken in \cite{zhou2017see} which contains an attention unit for weighted fusion of image embeddings. % in which the attention scores are \emph{learned implicitly} by final triplet loss. 
Score generation branch is designed for generating attention score in quality aware network \cite{liu2017qan}, region-based quality estimation network \cite{song2017region}, and diversity regularized spatiotemporal attention \cite{li2018diversity}. More recently, in order to take the parts into account, \cite{fu2018sta} proposed an attention mechanism by dividing the feature map into fixed set of horizontal parts.

%The generated attention score is also \emph{supervised implicitly}.

%\todo{use noindent+textbf to replace paragraph}
%\noindent\textbf{Attention-based Methods.}
%\eccvpara{}
%Some methods 
%\cite{wang2014person,huang2018video}
%are proposed to select the most discriminative video fragments to learn video representations, which can be regarded as \emph{fragment-level attention}. Recently, \emph{image-level attention} attracts a lot of interest and has been widely studied 
%\cite{zhou2017see,xu2017jointly,liu2017qan,song2017region,chen2018video,li2018diversity}.
%
%In Attentive Spatial-Temporal Pooling Networks (ASTPN) \cite{xu2017jointly} and Co-Attentive Embedding (CAE) \cite{chen2018video}, the computation of set representations in the gallery \emph{depends on the probe data}. They are \emph{less flexible and more computationally expensive} as the representations of sets in gallery need to be computed again when a different probe comes. 
%
%Temporal Attention Model (TAM) \cite{zhou2017see} contains an attention unit for weighted fusion of image representations, in which the attention scores are \emph{learned implicitly} by final triplet loss. 
%
%Score generation branch is designed for generating attention score in Quality Aware Network (QAN) \cite{liu2017qan}, Region-based Quality Estimation Network (RQEN) \cite{song2017region}, and Diversity Regularized Spatiotemporal Attention (DSRA) \cite{li2018diversity}. The generated attention score is also \emph{supervised implicitly}.

Our work has the following key differences: (1) we only consider spatial appearance information in the image set without using the temporal information and optical flow information as compared to  \cite{fu2018sta,zhu2016video,song2017region,li2018diversity}. (2) All quality-based approaches mentioned above learn the quality scores implicitly. In contrast, we learn quality with respect to set ID explicitly -- supervised by the ID information. (3) Unlike methods in \cite{liu2017qan,song2017region} which only deal with perceptual quality problem, our method can cope with a diverse set of quality problems. (4) ID-aware embedding does not depend on the probe data unlike the methods in \cite{xu2017jointly,chen2018video}, thus being more scalable and applicable. (5) Furthermore, although region-based feature extraction \cite{song2017region,li2018diversity,fu2018sta} has achieved great success, to demonstrate the robustness and effectivenss of IDE, we simply learn a global representation for every image in the set. \\

%In this work, we only consider rich spatial appearance information in the image set, without using the temporal information and optical flow following \cite{zhu2016video,song2017region,li2018diversity}. 
%
%ID-aware attention is \emph{supervised by the ID information explicitly} and covers diverse semantic quality problems beyond perceptual quality problems in QAN and RQEN. 
%ID-aware attention does not depend on the probe data, thus being flexible and efficient.
%
%Although region-based feature extraction \cite{song2017region,li2018diversity} has achieved great success, to demonstrate the robustness and effectivenss of IDAE, 
%we simply learn a global representation for every image in the set without designing the feature extraction module elaborately. 

\noindent
\textbf{Attention} Numerious works exist about attention, and they could be divided into two directions \cite{jetley2018learn,Andrea2019}. One direction is post hoc network analysis, and an attention unit in this direction relies on fully trained classification network model \cite{karen,Zhou_2016_CVPR}. The other one goes with trainable attention in which the weights of attention unit and the original network weights are learned jointly \cite{jetley2018learn,seo2016}. Our work aligns with the first direction which is the post hoc network analysis. However, our problem setting is verification, in which the label space between test and train data are disjoint. Thus the attention mechanism used during training cannot be used for testing.

%%%%%%%%%%%%%%%%%%%%%%%%%%%%%%%%%%%%%%%%%%%%%%%%%%%%%%%%%%
%\vspace{-4pt}
\section{Methodology}
\label{sec:proposed_method}
%%%%%%%%%%%%%%%%%%%%%%%%%%%%%%%%%%%%%%%%%%%%%%%%%%%%%
%\vspace{-2pt}
%\subsection{Overview} 
%Set-based person re-identification is a typical set-based verificcation problem \cite{li2018diversity,xie18a}. 
%The objective of SReID is to learn set embeddings and measure their similarity based on a metric. 
The overall pipeline of our framework, IDE, is shown in Fig. \ref{fig:overall_architecture}. First, the given image set, which consists of several person images, goes through CNN network and outputs representation for each image in the set. It is followed by a fully connected layer and softmax normalisation to generate ID-aware qualities. Then, these qualities are used in two ways: (1) Feature learning attention (FLA) component utilises the qualities such that medium hard images get more weights. This ends with the weighted cross-entropy loss for image-based classification. (2) Feature fusion attention (FFA) component transforms the qualites so that the ones with noise are given smaller weights to aggregate more discriminative information. This is supervised by the contrastive loss for set-based verification. IDE is trained end-to-end and optimized by two losses jointly. More formally, we aim to learn an embedding function $\Phi$ that takes an image set  $\mathbf{x} \in \mathbb{R}^{n\times h\times w\times c}$ as input, where $n$ is the number of images, $h,~w$ and $c$ are height, width and the number of channels respectively, and it outputs a $d$-dimensional discriminative feature vector $\mathbf{z}\in \mathbb{R}^{d}$, $\Phi: \mathbf{x} \rightarrow \mathbf{z}$. 

In what follows, we present the key components in detail, i.e., ID-aware quality generation, FLA, FFA, and loss functions. 

%To this end, as shown in Fig.~\ref{fig:overall_architecture}, we propose an end-to-end IDE framework, in which FLA aims to learn robust image embeddings and FFA targets fusing them well. 
%To learn robust image representations, FLA focuses on medium hard images whose classification confidences are around the centre of distribution, i.e., 0.5 as the distribution is between 0 and 1. This is a trade-off between gradient magnitude and gradient correctness. High quality images are easily classified and obtain high classification confidences (e.g., 0.99), but they are trivial because they contribute little to the loss and their gradients are really small. Low quality images (e.g. outliers) have low classification confidences (e.g., 0.01) and large gradients, but their gradient directions are misleading. 
%To fuse image features well, FFA emphasizes on high quality images to accumulate highly discriminative information into the set embedding.  Noticeably, high quality images are trivial in FLA but discriminative in FFA. Both FLA and FFA neglect low quality images.

%\vspace{-2pt}
\subsection{ID-aware Quality Generation}
\label{subsec: ID_aware}
ID-aware quality indicates how much it is semantically related to its set ID. We propose to generate it as follows. Firstly, we obtain image representations: for the $j$-th image set in the training batch,
$  (\mathbf{x}^j, y^j) = (\{\mathbf{x}^{j}_1, \mathbf{x}^{j}_2, ..., \mathbf{x}^{j}_n\}, y^{j}
) $, where $ \mathbf{x}^j_i \in \mathbb{R}^{h\times w\times c}$ and $y^j$ are an image and its identity, respectively (for brevity, hereafter we omit the superscript $j$ where appropriate),
we employ a deep CNN $f$ to embed each image $ \mathbf{x}_i$ to a $d$-dimensional feature representation, i.e, $\mathbf{z}_i = f(\mathbf{x}_i) \in \mathbb{R}^d$. Then, in order to obtain quality scores we apply a fully connected layer followed by a softmax normalisation. This way, 
the learned parameters $\{ \mathbf{c}_k \}^{C}_{k=1}$ of the fully  connected layer are ID context vectors, where $\mathbf{c}_k \in \mathbb{R}^d$ is the $k$-th ID's context vector  and $C$ is the number of identities in total. In short, $c_k$ plays a role of ID classifier.
% 
%We estimate one image's ID-aware quality by its classification confidence explicitly. 
%To obtain ID-aware qualities, there is a fully connected layer and a softmax normalization layer after the image representations. 
%
%The learned parameters $\{ \mathbf{c}_k \}^{C}_{k=1}$ of the fully  connected layer are ID context vectors, where $\mathbf{c}_k \in \mathbb{R}^d$ is the $k$-th ID's context vector  and $C$ is the number of identities.
%
More formally, the semantic relation of an image to an ID can be measured by the compatibility between the image's feature vector $\mathbf{z}_i$ and the ID's context vector $\mathbf{c}_k$. We calculate the dot product between two vectors to measure their compatibility \cite{jetley2018learn}, followed by a softmax operator which is used for normalising semantic relations over all identities. That is:
\begin{align}
	\label{equation:s_i}
	\mathrm{s}_i &= \frac{\exp(\mathbf{z}_i^\top\mathbf{c}_y)}{\sum\limits_{k=1}^C \exp(\mathbf{z}_i^\top\mathbf{c}_k)}.
\end{align}

\noindent
Inherently, Eq.~(\ref{equation:s_i}) computes the classification confidence (likelihood) of $\mathbf{x}_i$ with respect to its set ID $y$. Similarly, ID-aware quality is estimated by the classification confidence directly \cite{goldberger2016training}.%, which is similar to \cite{goldberger2016training} where the softmax output predicts the true label.    

%\vspace{-2pt}
\subsection{Feature Learning Attention}
\label{subsec: FLA}
%As shown in Fig.~\ref{fig:FLA_motivation}, 
During learning image embeddings, FLA focuses on medium hard images whose classification confidences are around the centre of distribution, i.e., 0.5, as the distribution is between 0 and 1. This is a trade-off between gradient magnitude and gradient correctness. Intuitively, high quality images are easily classified and obtain high classification confidences (e.g., 0.99), but they are trivial because they contribute little to the loss and their gradients are relatively small. Low quality images (e.g. outliers) have low classification confidences (e.g., 0.01) and large gradients, but their gradient directions are misleading. To achieve the trade-off between gradient magnitude and gradient correctness, medium hard images are given higher weights/attention in FLA. %The distribution of ID-aware of quality is between 0 and 1, i.e., $\mathrm{s}_i \in [0,1]$. 
The medium hard images are those whose ID-aware qualities are around the centre of distribution -- 0.5. 

\begin{figure}[!t]
	\centering
	\begin{subfigure}[b]{0.47\textwidth}
		\includegraphics[width=\textwidth]{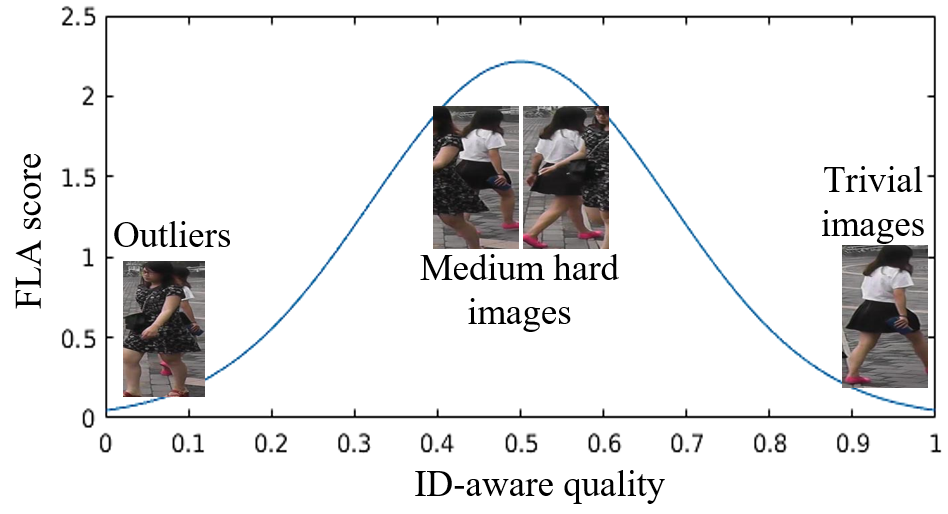}
		%\vspace{8pt}
	\end{subfigure}
	\caption{FLA aims to learn robust image-level embeddings. High quality images and low quality images are neglected because of small gradients and large but mis-leading gradients respectively; Medium hard images, i.e., medium quality images,  are emphasized, which is a trade-off between gradient magnitude and gradient correctness. }
	%\vspace{-14pt}
	\label{fig:FLA_motivation}
\end{figure}

To achieve this intution, we propose the following Gaussian function\footnote{Note that different functions could be also designed to reach the same goal. Exploring different functions are left for the future work.} to compute FLA scores from ID-aware qualities:     
\begin{align}
	\label{eq:fla}
	\mathrm{FLA}_i &= 
	%\frac{1}{{\sigma_\mathrm{FLA} \sqrt {2\pi } }} 
	\exp(
	{{{ - \left( {\mathrm{s}_i - 0.5 } \right)^2 } \mathord{\left/ {\vphantom {{ - \left( {\mathrm{s}(\mathrm{x}_i) - 0.5 } \right)^2 } {2 \sigma_\mathrm{FLA} ^2 }}} \right. \kern-\nulldelimiterspace} { \sigma_\mathrm{FLA} ^2 }}}
	),
\end{align}
where 0.5 is a mean, and $\sigma_\mathrm{FLA}$ is a temperature parameter controlling the distribution of FLA scores. $\mathrm{FLA}_i$ is the FLA score of $\mathbf{x}_i$, indicating its weight value  in the ID classification task. Note that a choice of 0.5 is due to our intution stated above. One intriguing property of this function is that we can change the distribution by merely changing the parameter $\sigma_\mathrm{FLA}$. Fig.~\ref{fig:FLA_motivation} illustrates FLA with Eq. (\ref{eq:fla}).   %We illustrate this in   Fig. \ref{}.

%%%%%%%%%%%
%%%%%%%%%%

%%%%%%%%%%%%%%%%%%%%%%%%%%%%
%\vspace{-2pt}
\subsection{Feature Fusion Attention}
\begin{figure}[!t]
	\centering
	\begin{subfigure}[b]{0.47\textwidth}
		\includegraphics[width=\textwidth]{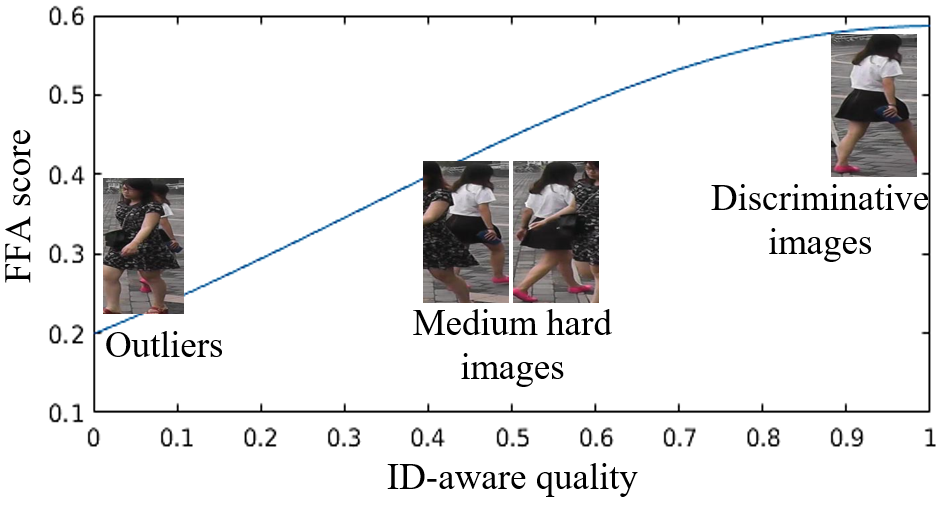}
	\end{subfigure}
	%\vspace{-24pt}
	\caption{
		FFA fuses image-level representations to obtain the set-level embedding. More discriminative images in the set are emphasized to accumulate more discriminative information into the set representation. The images with higher ID-aware quality are more discriminative.
	}
	%\label{fig:motivation}
	\label{fig:FFA_motivation}
	%\vspace{-14pt}
\end{figure}

During feature fusion which is used for verification in a later stage, FFA aggregates most informative feature embeddings. In general, high quality images (ID-aware quality) are very discriminative, thus being more informative for embedding the set. %Therefore, FFA attends more attention to the discriminative images which have high ID-aware quality. 
This is shown in Fig.~\ref{fig:FFA_motivation}. To realise this intuition, we compute FFA scores from ID-aware qualities using the following Gaussian function with $\mu=1$:   
\begin{align}
	\mathrm{FFA}_i &= 
	%\frac{1}{{\sigma_\mathrm{FFA} \sqrt {2\pi } }} 
	\exp(
	{{{ - \left( {\mathrm{s}_i - 1 } \right)^2 } \mathord{\left/ {\vphantom {{ - \left( {\mathrm{s}(\mathrm{x}_i) - 1 } \right)^2 } {2 \sigma_\mathrm{FFA} ^2 }}} \right. \kern-\nulldelimiterspace} { \sigma_\mathrm{FFA} ^2 }}}
	),
\end{align}
where $\sigma_\mathrm{FFA}$ is a temperature parameter controlling the distribution of FFA scores. 
$\mathrm{FFA}_i$ is the score of $\mathbf{x}_i$, indicating its importance when being aggregated into the set embedding.  
As a result, we obtain the set representation by weighted fusion of image representations in the set:
\begin{equation}
	\Phi\left(\mathbf{x} \right) = \frac{
		\sum\limits_{i=1}^n{ \mathrm{FFA}_i  \cdot \mathbf{z}_i  }
	}
	{
		\sum\limits_{i=1}^n{ \mathrm{FFA}_i }
	}.
\end{equation}
The term in denominator $\sum\limits_{i=1}^n{ \mathrm{FFA}_i }$ is the sum of FFA scores for normalisation. At each iteration, FFA scores of images are computed in the forward process and used as constant values for just scaling the gradient vectors during gradient back-propagation. 
%\vspace{-2pt}

%%%%%%%%%%%
%%%%%%%%%%

%\vspace{-2pt}
\subsection{Loss Functions}
Suppose there are $m$ image sets in each mini-batch, i.e.,$\{ (\mathbf{x}^j, y^j) \}^{m}_{j=1}$ and 
each set contains $n$ images, thus the mini-batch size is $mn$.\\

\noindent
\textbf{Weighted Cross-Entropy Loss}.
To learn robust image representations based on FLA scores, we propose a weighted cross-entropy loss for the image-level ID classification task:
\begin{equation}
	L_{\mathrm{WCEL}} = - \frac{ 
		\sum\limits_{j=1}^{m}
		\sum\limits_{i=1}^{n}
		({\mathrm{FLA}^j_i \cdot \log{\mathrm{s}^j_i}}
		)
	}
	{
		\sum\limits_{j=1}^{m}
		\sum\limits_{i=1}^{n}
		{\mathrm{FLA}^j_i }
	},
\end{equation}
where the term in the denominator $\sum\limits_{j=1}^{m}
\sum\limits_{i=1}^{n}
{\mathrm{FLA}^j_i }$ is for normalisation. 
Accordingly, the partial derivative of $L_{\mathrm{WCEL}}$ w.r.t. $\mathrm{s}^j_i$ is: 
\begin{align}
	\label{equation: derivative}
	\frac{
		\partial L_{\mathrm{WCEL}}
	}
	{
		\partial {\mathrm{s}^j_i}
	} 
	= & 
	\frac{\mathrm{FLA}^j_i} {{\sum\limits_{j=1}^{m}\sum\limits_{i=1}^n {\mathrm{FLA}^j_i}}}
	\cdot
	(- \frac{1} {\mathrm{s}^j_i}). 	  
\end{align}
At each iteration, after being computed in the forward process, FLA scores of images are assumed as constant values for scaling the gradients during the  back-propagation process. Compared with standard cross-entropy loss, we use the normalised FLA score $\frac{\mathrm{FLA}^j_i} {{\sum\limits_{j=1}^{m}\sum\limits_{i=1}^n {\mathrm{FLA}^j_i}}}$ to scale image's gradient.\\ 

\noindent
\textbf{Contrastive Loss}. For set-based verification, on top of set-level representations, we employ contrastive loss \cite{hadsell2006dimensionality} using the multi-batch setting \cite{tadmor2016learning}. The contrastive loss is a well-known loss function used for verification task. It pulls sets from the same identity as close as possible and pushes the sets from different identities farther than a pre-defined margin $\alpha$. Specifically, we construct a pair between every two set embeddings, resulting in $m(m-1)/2$ pairs in total.
For all the set embeddings $\{ (\mathbf{z}^j, y^j) \}^{m}_{j=1}$ in the mini-batch, we compute the contrastive loss per pair:
\begin{equation}
	L(\alpha; \mathbf{\Phi}^{j}, \mathbf{\Phi}^{k}, y^{jk}) = y^{jk}d_{jk}^2 + (1-y^{jk})\max(0, \alpha - d_{jk})^2,
\end{equation}
where $y^{jk}=1$ if $y^{j}=y^{k}$ and $y^{jk}=0$, otherwise. 
%if $y^{j} \neq y^{k}$. 
$d_{jk} = \left\Vert
\mathbf{\Phi}^j - \mathbf{\Phi}^k
\right\Vert_2 $  
is the distance between the set pair. The contrastive loss of the mini-batch is the average loss over all the pairs:
\begin{equation}
	\label{equation: CL}
	L_{\mathrm{CL}} = 
	\frac{2}{m(m-1)}
	\sum_{j=1}^{m-1}
	\sum_{k=j+1}^{m}
	L(\alpha; \mathbf{\Phi}^{j}, \mathbf{\Phi}^{k}, y^{jk}).
\end{equation}
\noindent
IDE is trained end-to-end by optimising the weighted cross-entropy loss and set-based contrastive loss jointly:
\begin{equation}
	L = 
	L_{\mathrm{WCEL}}
	+
	L_{\mathrm{CL}}.
\end{equation}
%\subsection{Matching for Set-based Verification}
After the training is finished, the trained CNN model can be applied to extract image features. Generally, the label spaces of the training and testing sets are disjoint in the verification problems. Therefore, we cannot estimate the ID-aware quality of testing images as their classification confidences are not available during testing. 
To obtain the set embedding in the test phase, we aggregate image representations in the set simply by average fusion. After that, we verify whether two image sets show the same person simply by computing their cosine distance. \\

\noindent
\textbf{Remarks:} 1) We would like to note that although we use the term `attention' to mean the weight for images, our method is \textit{different} from the classical attention-based methods. In our formulation, the scores of FLA and FFA are taken as a constant values after forward propogation. In contrast, standard attention-based methods would allow the gradients to flow through FFA and FLA. 2) During training we can estimate the quality, however we cannot estimate the quality at the testing stage. Our main goal during training is to learn a robust embedding function by ignoring perceptually and semantically low quality images. This way we assume that our model generalises better because it does not overfit to the training dataset. Our extensive experiments validate that this assumption is reasonable. Please see Sec. \ref{related_work} for more. \\
\section{Experiments}
\label{sec:experiments}

%%%%%%%%%%%%%%%%%%%%%%%%%%%%%%%%%%%%%%%%%%%%%%%%%%
\subsection{Datasets and Settings}

%\textbf{Datasets. } 
\noindent\textbf{Datasets.} 
We use two large-scale and two small-scale datasets in our experiments:  
(1) \textit{LPW} \cite{song2017region} is a large-scale dataset released recently. The persons in the dataset are collected across three different scenes separately. Three cameras are placed in the first scene, while four cameras are placed in other two scenes. Each person is captured by more than one camera so that cross-camera search could be possible in each scene. There are totally 7694 image sets with about 77 images per set. Following the evaluation setting in \cite{song2017region}, 1975 persons captured in the 2$^{nd}$ scene and 3$^{rd}$ scene are used for training, while 756 persons from the first scene are used for testing. The dataset is challenging as the evaluation protocol is close to real-world situation, that is, the training data and the testing data are different in terms of not only identities but also scenes. (2) \textit{MARS} \cite{zheng2016mars} is another large-scale dataset. There are 20478 tracklets (image sets) of 1261 persons in total and each person is shot by at least two cameras. This dataset is challenging due to automatic detection and tracking errors. (3) \textit{iLIDS-VID} \cite{wang2014person} is a small-scale dataset. Since it is collected in airport environment, image sequences contain significant viewpoint variations, occlusions and background clutter. There are 300 identities in total and each identity has two tracklets from two different cameras. (4) \textit{PRID2011} \cite{hirzer2011person} consists of 400 tracklets for 200 persons from two cameras. The tracklet length varies from 5 to 675. Compared to the previous datasets, PRID2011 is less challenging because of few variations and rare occlusion. For LPW and MARS, we follow the evaluation setting in \cite{song2017region} and \cite{zheng2016mars} respectively. For iLIDS-VID and PRID2011, they are split into two subsets with equal size following \cite{wang2014person}, one for training and the other for testing. In addition, the 10 random trials are fixed and the same as in \cite{wang2014person} for fair comparison. A summary of the datasets is shown in Table \ref{table:dataset}. 
%%%%%%
\setlength{\tabcolsep}{2pt}
\begin{table}[t!]
	\begin{center}
		\caption{A summary of four set-based person ReID datasets used in our experiments.  %(`\#` indicates the number of items).
		}
		%\tabcapspace
		\label{table:dataset}
		%\vspace{-8pt}
		\begin{tabular}{lllll}
			\hline\noalign{\smallskip}
			Datasets & LPW & MARS & iLIDS-VID & PRID2011 \\ 
			\noalign{\smallskip}
			\hline
			\noalign{\smallskip}
			
			\#identities & 2731 & 1261 & 300 & 200 \\
			\#boxes & 590,547 & 1,067,516 & 43,800 &  40,000 \\
			\#tracklets & 7694 & 20,715 & 600 & 400 \\
			\#cameras & 11 & 6 & 2 & 2\\
			
			DT failure & No & Yes & No & No \\
			resolution & 256x128 & 256x128 & 128x64 & 128x64 \\
			detection & detector & detector & hand & hand \\
			evaluation & CMC & CMC \& mAP & CMC & CMC\\
			\hline
		\end{tabular}
		%\vspace{-16pt}
	\end{center}
\end{table}
%\tabtextspace
%
%%%%%

\noindent\textbf{Evaluation Metrics.}
%\eccvpara{Evaluation Metrics.}
We report the Cumulated Matching Characteristics (CMC) results for all the datasets. We also report the mean average precision (mAP) for MARS following the common practice.    

\noindent \textbf{Implementation Details.}
%\eccvpara{Implementation Details.}
We use GoogLeNet with batch normalisation \cite{ioffe2015batch} as our backbone architecture. Every input image is resized to $224\times224$. We do not apply any data augmentation for training and testing.
Each mini-batch contains 3 persons, 2 image sets per person, 9 images per set, so the batch size is 54 ($m = 6, n = 9$).  
Each image set in the mini-batch is randomly sampled from the complete image set during training.
According to Eq.~(\ref{equation: CL}), we have 3 positive set pairs and 12 negative set pairs totally, thus the positive-to-negative rate is 1:4.
The margin of contrastive loss is set to 1.2 ($\alpha=1.2$). 
Stochastic gradient descent (SGD) optimiser is applied with an initial learning rate of $1e^{-3}$. 
When training the model on each dataset, we initialise it by the pre-trained GoogLeNet model on ImageNet.
%
%We train our model in a two-stage way. Firstly, we train an IDE (ID embedding) model \cite{zheng2017person} with standard softmax loss. Then, we train our model initialized by the IDE model. We report the results of IDE and our model in the experiment section. The reason why we train our model stage by stage is that the IDE model has been shown to extract discriminative features \cite{zheng2016mars}, thus offers a reasonable initialization for our ID-aware attention units in both tasks. 
%
We use Caffe \cite{jia2014caffe} for implementation. 

For the temperature parameters $\sigma_\mathrm{FLA}$ and $\sigma_\mathrm{FFA}$, they control the variances of FLA scores and FFA scores respectively. Based on our experimental results, they are insensitive and are fixed in all the experiments ($\sigma_\mathrm{FLA} = 0.18, \sigma_\mathrm{FFA} = 0.68 $) although better results could be obtained by exploring optimal parameters for each dataset. Analysis of these parameters is reported in the supplimentary material. 

%%%%%%%%%%%%%%%%%%%%%%%%%%%%
\setlength{\tabcolsep}{2pt}
\begin{table}[t!]
	\begin{center}
		\caption{Comparison with state-of-the-art methods on MARS in terms of CMC-1(\%) and mAP(\%). Quality-based methods are indicated by `Quality' column. \textcolor{blue}{Blue} indicates `re-ranking (RR)'. 
			`Custom' means the backbone is customised.}
		%\tabcapspace
		%\vspace{-8pt}
		\label{table:literature_mars}
		\begin{tabular}{lccccc}
			\hline\noalign{\smallskip}
			Methods & Quality & Backbone &   CMC-1  & mAP\\
			\noalign{\smallskip}
			\hline
			IDE+Euc  & No & CaffeNet & 58.7 & 40.4 \\
			IDE+XQDA  & No & CaffeNet & 65.3 & 47.6 \\
			IDE+XQDA+RR  & No & CaffeNet & \textcolor{blue}{67.8} &  \textcolor{blue}{58.0} \\
			IDE  & No & ResNet50 & 62.7 &  44.1 \\
			IDE+XQDA  & No & ResNet50 & 70.5 &  55.1 \\
			IDE+XQDA+RR  & No & ResNet50 & \textcolor{blue}{73.9} &  \textcolor{blue}{68.5} \\
			CNN+RNN         &  No & Custom & 43.0 & --\\
			CNN+RNN+XQDA  & No & Custom & 52.0 & --\\
			AMOC+EpicFlow  & No & Custom & 68.3 & 52.9 \\
			%MSCAN \cite{li2017learning} & 71.8 & 86.6 &  56.1 & --\\
			%TriNet \cite{hermans2017defense} & 79.8 & 91.4 & -- & 67.7 \\
			%TriNet+Re-ranking \cite{hermans2017defense} & 81.2 & 90.8 & -- & 77.4 \\
			\hline
			ASTPN           & Yes & Custom & 44.0 & -- \\
			SRM+TAM  & Yes & CaffeNet & 70.6 & 50.7 \\
			RQEN  & Yes & GoogleNet & 73.7 & 51.7 \\
			RQEN+XQDA+RR & Yes & GoogleNet & \textcolor{blue}{77.8} & \textcolor{blue}{71.1} \\
			DRSA & Yes & ResNet50 & 82.3 & 65.8 \\
			CAE & Yes & ResNet50 & 82.4 & 67.5 \\
			\hline
			\textbf{Ours}  & Yes & GoogleNet & \textbf{83.3}  & \textbf{71.7}  \\
			\textbf{Ours}+RR  & Yes & GoogleNet & \textcolor{blue}{\textbf{85.1}}  & \textcolor{blue}{\textbf{82.2}}  \\
			\hline
		\end{tabular}
		%\vspace{-16pt}
	\end{center}
\end{table}
%\tabtextspace
%%%%%%%%%%%%%%%%%%%%%%%%%%%%

%%%%%%%%%%%%%%%%%%%%%
\setlength{\tabcolsep}{1.4pt}
\begin{table}[t!]
	\centering
	\caption{Comparison with state-of-the-art methods on iLIDS-VID, PRID2011 and LPW in terms of CMC-1(\%). Quality-based methods are indicated by `Quality' column.
		LPW is a difficult cross-scene dataset and released in \cite{song2017region} recently.}
	%\tabcapspace
	%\vspace{-8pt}
	\label{table:literature_ilid_prid}
	\begin{tabular}{lcccc}
		\hline
		Method           & Quality & iLIDS-VID & PRID2011 & LPW  \\ 
		\hline
		STA              & No        & 44.3      & 64.1     & --     \\ 
		SI$^2$DL  & No & 48.7 & 76.7 & -- \\
		IDE+XQDA   & No        & 53.0      & 77.3     & --     \\ 
		TDL              & No        & 56.3      & 56.7     & --     \\ 
		
		CNN+RNN          & No        & 58.0      & 70.0     & --     \\ 
		AMOC+EpicFlow  & No & 68.7 & 83.7 & -- \\
		\hline
		ASTPN            & Yes       & 62.0      & 77.0     & --     \\ 
		SRM+TAM  & Yes       & 55.2      & 79.4     & --     \\ 
		QAN               & Yes       & 68.0      & 90.3     & --     \\ 
		RQEN    & Yes       & 76.1      & 92.4     & 57.1 \\ 
		DRSA    & Yes       & 80.2      & 93.2     & -- \\
		CAE & Yes &  74.4 & 86.0 &-- \\ 
		\hline
		\textbf{Ours}             & Yes       & \textbf{81.9}      & \textbf{93.7}     & \textbf{70.9} \\
		\hline
	\end{tabular}
	%\vspace{-16pt}
\end{table}
%\tabtextspace
%%%%%%%%%%%%%%%

%%%%%%%%%%%%%%%%%%%%%%%%%%%%%%%%%%
\subsection{Comparison with State-of-the-art Methods}
\label{subsec:state_of_the_art}
%\vspace{2pt}
\noindent\textbf{MARS Dataset.} 
%\noindent\textbf{MARS Dataset.} 
We compare our approach with several state-of-the-art methods. Overall, we divide them into two categories: non-quality-based and attention-based methods. 
\emph{The non-quality-based methods}: 
the softmax loss is used during training with custom/common networks, e.g., IDE(CaffeNet) \cite{zheng2016mars}, IDE(ResNet50) \cite{zhong2017re}, 
and metric learning/re-ranking strategies are applied for boosting the performance, e.g., IDE(CaffeNet)+XQDA \cite{zheng2016mars}, IDE(CaffeNet)+XQDA+RR \cite{zhong2017re}, IDE(ResNet50)+XQDA \cite{zhong2017re}, IDE(ResNet50)+XQDA+RR \cite{zhong2017re}; 
the temporal information of image sequences is used by combining recurrent neural networks with convolutional networks (CNN+RNN \cite{mclaughlin2017video}) and again metric learning is used on top of this (CNN+RNN+XQDA); 
optical flow is used to capture motion information (AMOC+EpicFlow \cite{liu2017video}). 
\emph{The quality-based methods}: combining image-level attention and temporal pooling to select informative images (ASTPN \cite{xu2017jointly}, SRM+TAM \cite{zhou2017see}, CAE\footnote{For CAE, we present the results of complete sequence instead of multiple subsets so that it can be compared with other methods. Multiple subsets can be regarded as data augmentation.} \cite{chen2018video}); designing attentive spatiotemporal models to extract complementary region-based information between images in the set (RQEN \cite{song2017region}, DRSA\cite{li2018diversity}). 
%\textit{Results.} 
%\eccvpara{Results.} 

Results of our method and compared methods are shown in Table \ref{table:literature_mars}. \emph{The following key findings are observed}: 
\emph{(1)} Our method considerably outperforms all the compared methods in both metrics (CMC, mAP). Noticeably,  our mAP is around 6\% better than DSRA and 4\% better than CAE. DSRA applies spatiotemporal attention to extract features of latent regions  and is pre-trained on several large image-based person ReID datasets. CAE is a co-attentive embedding model and computationally expensive as the embeddings of sets in the gallery are dependent on the probe data. In addition, CAE combines optical flow information and RGB information.
\emph{(2)} Generally, the quality-based methods outperform non-quality-based methods except for ASTPN with a relatively shallow architecture. 
\emph{(3)} When these methods are coupled with re-ranking technique, there is always a performance boost (ours achieves 82.2 \% in terms of mAP). 

%\vspace{2pt}
\noindent\textbf{iLIDS-VID, PRID2011, and LPW datasets.} 
%\eccvpara{iLIDS-VID, PRID2011, and LPW datasets.} 
Grouping the methods for these datasets is the same, and most compared methods are borrowed from Table \ref{table:literature_mars}. Additional methods include: a spatial-temporal body-action model (STA \cite{liu2015spatio}), novel metric learning approaches (SI$^2$DL \cite{zhu2016video}, TDL \cite{you2016top}). 
%\eccvpara{Results.} 

The results are shown in Table~\ref{table:literature_ilid_prid}. We find that our observations are consistent with MARS dataset. In particular: 
\emph{(1)} In all datasets, our method achieves the best performance; 
\emph{(2)} The margin of iLIDS-VID and LPW  is larger than PRID2011. This is because PRID2011 is a much cleaner dataset and its accuracy is already high; 
\emph{(3)} Noticeably the LPW dataset is the most challenging since the scenes are different for training and testing, and yet our method is around 14\% higher than RQEN \cite{song2017region}, showing the generalisation capability of our proposed method.

%%%%%%%%%%%%%
\setlength{\tabcolsep}{6pt}
\begin{table}[t!]
	\begin{center}
		\caption{Effectiveness of FLA and FFA. The ablation studies are conducted on large dataset LPW. CMC (\%) results are presented. 
			Baseline means standard cross-entropy loss and average fusion are applied. 
			Compared with the baseline, FLA applies weighted cross-entropy loss while FFA uses weighted fusion. 
			FLA+FFA combines weighted cross-entropy loss and weighted fusion. 
		}
		%\tabcapspace
		%\vspace{-8pt}
		\label{table:component_effectiveness_v1}
		\begin{tabular}{lcccc}
			\hline
			Model &  1 & 5 & 10 & 20 \\
			\hline
			%			Baseline & High-level & 56.3 & 81.3 & 89.0 & 92.6 \\
			%			FFA & High-level & 64.6 & 85.3 & 90.6 & 94.6 \\
			%			FLA & High-level & 63.2 & 86.0 & \textbf{91.8} & \textbf{95.4} \\
			%			FLA+FFA & High-level & \textbf{65.5} & \textbf{87.0} & 91.4 & 94.2 \\
			%			\hline
			Baseline &  63.2 & 85.4 & 90.6 & 94.8 \\
			FFA & 67.9 & 88.8 & 93.3 & 95.8 \\
			FLA & 68.8 & 88.1 & 93.5 & \textbf{96.3} \\
			FFA+FLA &  \textbf{70.9} & \textbf{89.3} & \textbf{93.9} & 95.8 \\
			\hline
		\end{tabular}
		%\vspace{-16pt}
	\end{center}
\end{table}
%\tabtextspace
%%%%%%%%%%%%%%%

%%%%%%%%%%%%%%%%%%%%%%%%%%%%%%%%%%%%%%%%%%%%%%%%%%%%%%%%%%%%
%\vspace{-2pt}
\subsection{Effectiveness of FLA and FFA}
\label{sec: two_attention_unit}

IDE has two key components: (1) FLA aims to learn robust image representations. Based on FLA, weighted cross-entropy loss is proposed to replace standard cross-entropy loss in the baseline; (2) During training, FFA is proposed for weighted fusion of image embeddings to replace average fusion in the baseline. We use LPW for this analysis. Table \ref{table:component_effectiveness_v1} shows that the performance improves significantly using either FLA or FFA compared to the baseline, demonstrating the effectiveness of FLA and FFA. We obtain the best performance by FFA+FLA at ranks ranging from 1 to 10.% which are more important than rank at 20. 
%The ablation studies are conducted on LPW using two different feature representations: high-level representations from $pool5/{7\times7}\_s1$ layer, and mid-level representations from $loss2/f c/bn/sc/relu$ layer. 
%The dimension of features is 1024.
%We find that for every model, the performance of mid-level features is significantly better than high-level features. 
%Mid-level representation contains rich context information and captures appearance information and latent attribute information well \cite{hariharan2015hypercolumns,diba2016deepcamp,vittayakorn2016automatic,he2017mask,bau2017network}.
%Hence, we choose the mid-level representation as the final embedding.

\subsection{A Variant of FFA}
\label{sec: variant_ffa}
%%%%%%
We employ the same idea as FLA which pays more attention to medium hard samples to see the impact. We name this as FFA\textsubscript{MH}. Table~\ref{table:different_fusion} shows that in both settings (without FLA or with FLA) FFA\textsubscript{MH} performs similar as average fusion but much worse than our proposed FFA. This validates our assumption that assigning more weight to high quality images is desirable at the verification stage.
%%%%%

%
%%%%%%
%\vspace{-12pt}
\setlength{\tabcolsep}{10pt}
\begin{table}
	\begin{center}
		\caption{
			Evaluation of FFA\textsubscript{MH} on LPW
			in terms of CMC (\%). 
			FFA\textsubscript{MH} is a variant of FFA and attends more attention to medium hard images.
			The results of Baseline (average fusion), FLA(average fusion), FFA, and FLA+FFA are copied from Table~\ref{table:component_effectiveness_v1}.
			We compare them in two different settings: without FLA, and with FLA.}
		\label{table:different_fusion}
		%\vspace{-8pt}
		\begin{tabular}{lcccc}
			\hline
			Model  & 1 & 5 & 10 & 20 \\
			\hline
			Baseline & 63.2 & 85.4 & 90.6 & 94.8 \\
			FFA &  \textbf{67.9} & \textbf{88.8} & \textbf{93.3} & \textbf{95.8} \\
			FFA\textsubscript{MH} & 64.5 & 86.8 & 91.5 & 94.3 \\
			\hline
			FLA & 68.8 & 88.1 & 93.5 & \textbf{96.3} \\
			FLA+FFA & \textbf{70.9} & \textbf{89.3} & \textbf{93.9} & 95.8 \\
			FLA+FFA\textsubscript{MH} & 68.5 & 89.2 & 92.7 & 95.8 \\
			\hline
		\end{tabular}
	\end{center}
	%\vspace{-12pt}
\end{table}
%%%%%

%%%%%%%%%%%%%%%%%%%%%%%%%%%%%%%%%%%%%%%%%%%%%%%%%%%%%
%\vspace{-2pt}
\subsection{Cross-Dataset Evaluation}

Cross-dataset testing is a better way to evaluate a system's real-world performance than only evaluating its performance on the same dataset used for training. Generally any public dataset only represents a small proportion of all real-world data. The model trained on A dataset could perform much worse when applied to B dataset, which indicates the model overfits to the particular scenario. 

We conduct cross-dataset testing on PRID2011 to evaluate the generalisation of our method. The diverse and large MARS and iLIDS-VID are used for training. 
We follow the evaluation setting in CNN-RNN \cite{mclaughlin2017video} and ASTPN \cite{xu2017jointly}, i.e., the model is tested on 50\% of PRID2011. We use the same testing data as Table~\ref{table:literature_ilid_prid} so the cross-dataset testing results can be compared with the results in Table~\ref{table:literature_ilid_prid}. 
The results are shown in Table~\ref{table:cross_dataset_test}.\footnote{Cross-dataset experimental results of QAN and RQEN are not reported as they were tested on 100\% of PRID}%, thus cannot be compared with results in Table~\ref{table:literature_ilid_prid}
%	and Table~\ref{table:cross_dataset_test}.}  
%
As expected, the results are worse than within-dataset testing because of dataset bias.
However, our method generalises well and achieves state-of-the-art cross-dataset testing performance.
For CMC-1 accuracy, our method achieves 41.4\% when trained on MARS and 61.7\% when trained on iLIDS-VID. Noticeably, it is comparable with the performance (64.1\%) of spatial-temporal body-action model STA \cite{liu2015spatio} in Table~\ref{table:literature_ilid_prid}.
This indicates that IDE enhances the generalisation ability significantly. \\

%%%%%%%%%%%%%%%%%%%%%%%%%%%%
\setlength{\tabcolsep}{2pt}
\begin{table}
	\begin{center}
		\caption{Cross-dataset testing in terms of CMC (\%). }
		\label{table:cross_dataset_test}
		%\vspace{-8pt}
		\begin{tabular}{lcccccc}
			\hline
			Method & Train & Test & 1 & 5 & 10 & 20  \\
			\hline
			%\noalign{\smallskip}
			CNN+Euc  & MARS & PRID2011 & 7.6 & 24.6 & 39.0 & 51.8 \\
			CNN+RNN  & MARS & PRID2011  & 18.0 & 46.0 & 61.0 & 74.0 \\
			Ours & MARS & PRID2011 & \textbf{41.4}  & \textbf{57.5} & \textbf{69.1} & \textbf{81.2}  \\
			\hline
			CNN+RNN  & iLIDS-VID & PRID2011  & 28.0 & 57.0 & 69.0 & 81.0 \\
			ASTPN  & iLIDS-VID & PRID2011  & 30.0 & 58.0 & 71.0 & 85.0 \\
			Ours & iLIDS-VID & PRID2011 & \textbf{61.7}  & \textbf{88.5} & \textbf{95.4} & \textbf{98.8} \\
			\hline
		\end{tabular}
		%\vspace{-16pt}
		\vspace{-0.2cm}
	\end{center}
\end{table}

%%%%%%%%%%%%%%%%

%%%%%%%%%%%%%%%%%%%%%%%%%%%%%%%%%%%%%%%%%%%%%%%%%%%%%%%%%%
%\vspace{-4pt}
\vspace{-0.3cm}
\section{Conclusion}

In this paper we propose a new concept, ID-aware  quality, which measures the semantic quality of images guided by their ID information. 
Based on ID-aware quality, we propose ID-aware embedding which contains FLA and FFA. 
To learn robust image embeddings, FLA gives more weight to medium hard images.
To accumulate more discriminative information when fusing image features, FFA assigns more weight to high quality images.
Compared with previous state-of-the-art attentive spatiotemporal methods, our method works much better in both within-dataset testing and cross-dataset testing. Furthermore, IDE is much simpler compared with previous attentive spatiotemporal methods.   

% In our future work, we will explore combining our image-level attention units in videos with region-level attention in images.   
%%%%%%%%%%%%%%%%%%%%%%%%%%%%%%%%%

%start-------------------------------------------------------------------------

%end-------------------------------------------------------------------------

%%%%%%%%%%%%%%%%%%%%%%%%%%%%%%%%%%%
%%REFERENCES
%%%%%%%%%%%%%%%%%%%%%%%%%%%%%%%%%%%
{\small
	\bibliographystyle{ieee}
	\bibliography{references_aaai2019.bib}
}

\newpage

%\title{ID-aware Quality for Set-based Person Re-identification \\Supplementary Material: Hyperparameters Analysis 
%	}
%
%\maketitle
%\setcounter{equation}{24}
%\setcounter{table}{3}
%\setcounter{figure}{4}
\setcounter{section}{0}

%start-------------------------------------------------------------------------
%\vspace{-2pt}
\section{The temperature parameter of FLA: $\sigma_{\mathrm{FLA}}$ }

\begin{figure}[!h]
	\centering
	\includegraphics[width=1.0\linewidth]{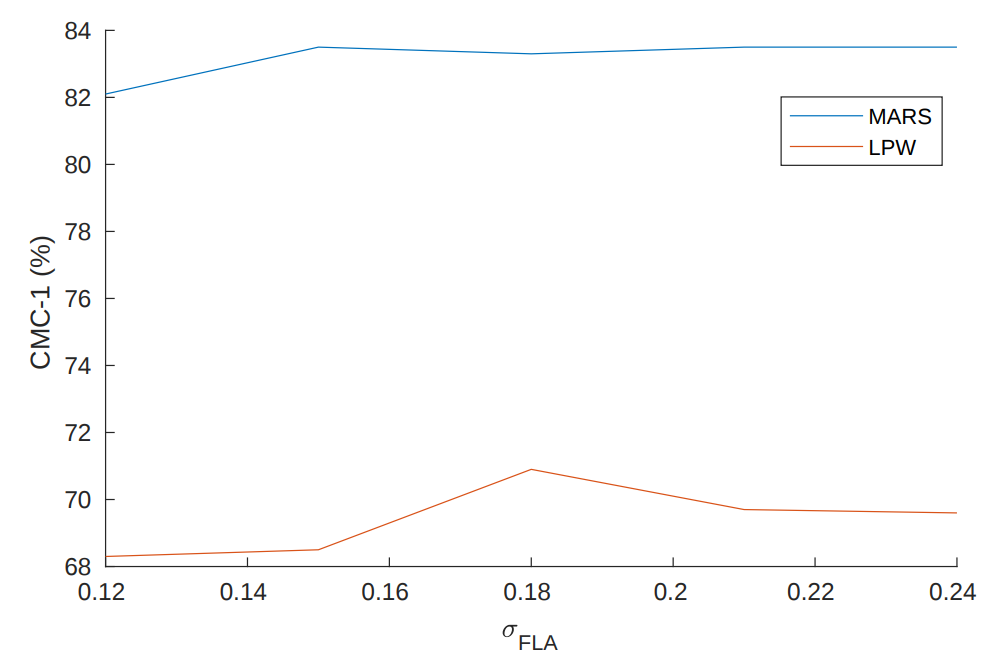}	
	\caption{
		The results of different $\sigma_{\mathrm{FLA}}$ on MARS and LPW in terms of CMC-1 (\%).  
		This is a line figure representation of results shown in Table~\ref{table:different_sigma_FLA}.
	}
	\label{fig:different_sigma_FLA}
	\vspace{-25pt}
\end{figure}

To study the impact of $\sigma_{\mathrm{FLA}}$, we set the temperature of FFA $\sigma_{\mathrm{FFA}}=0.68$ in all experiments. We conduct experiments on MARS and LPW and report the results in Table~\ref{table:different_sigma_FLA} and Figure~\ref{fig:different_sigma_FLA}. 
Firstly, we can see that the performance is not sensitive to $\sigma_{\mathrm{FLA}}$. For example, the performance difference is smaller than 1.5\% on MARS when $\sigma_{\mathrm{FLA}}$ ranges from 0.12 to 0.24.  
Secondly, better results can be obtained by exploring the optimal $\sigma_{\mathrm{FLA}}$ on each dataset. In the main paper, we fix $\sigma_{\mathrm{FLA}}=0.18$ on all datasets. We observe that $\sigma_{\mathrm{FLA}}=0.18$ works best on LPW but not on MARS.

%%%%%
\begin{table}[!h]
	\begin{center}
		\caption{
			The results of different $\sigma_{\mathrm{FLA}}$ on MARS and LPW in terms of CMC-1 (\%).
			We fix $\sigma_{\mathrm{FFA}}=0.68$ in all experiments. 
		}
		\label{table:different_sigma_FLA}
		\vspace{-8pt}
		\begin{tabular}{lcc}
			\hline
			$\sigma_{\mathrm{FFA}}=0.68$  & MARS & LPW \\
			\hline
			$\sigma_{\mathrm{FLA}}=0.12$ & 82.1 & 68.3 \\
			$\sigma_{\mathrm{FLA}}=0.15$ & 83.5 & 68.5 \\
			$\sigma_{\mathrm{FLA}}=0.18$ & 83.3 & 70.9 \\
			$\sigma_{\mathrm{FLA}}=0.21$ & 83.5 & 69.7 \\
			$\sigma_{\mathrm{FLA}}=0.24$ & 83.5 & 69.6 \\
			\hline
		\end{tabular}
	\end{center}
\end{table}
%%%%%

\section{The temperature parameter of FFA: $\sigma_{\mathrm{FFA}}$ }

\begin{figure}[!h]
	\centering
	\includegraphics[width=1.0\linewidth]{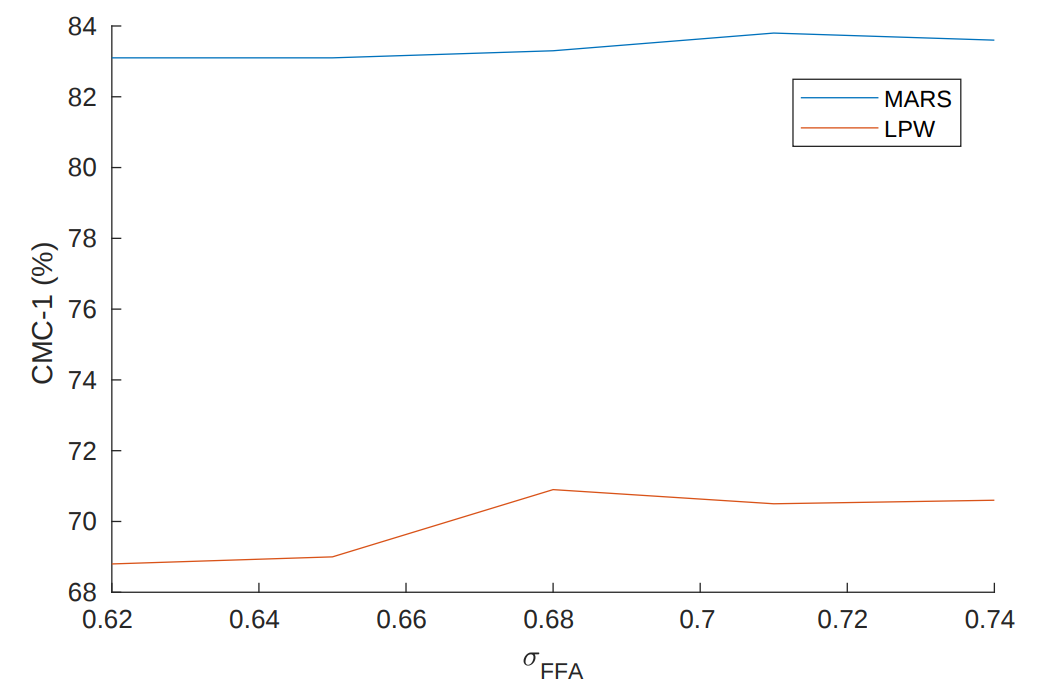}
	\caption{
		The results of different $\sigma_{\mathrm{FFA}}$ on MARS and LPW in terms of CMC-1 (\%).
		This is the line figure illustration of results presented in Table~\ref{table:different_sigma_FFA}.
	}
	\vspace{-25pt}
	\label{fig:different_sigma_FFA}
\end{figure}

To study the influence of $\sigma_{\mathrm{FFA}}$, we set $\sigma_{\mathrm{FLA}}=0.18$ in all experiments. The experiments are conducted on MARS and LPW and their results are presented in Table~\ref{table:different_sigma_FFA} and Figure~\ref{fig:different_sigma_FFA}. 
First, we observe that the performance is also insensitive to $\sigma_{\mathrm{FFA}}$.  The performance gap is less than 1.0\% on MARS and around 2.0\% on LPW.
Second, we can also obtain better results by searching optimal parameters for different datasets. 
We fix $\sigma_{\mathrm{FFA}}=0.68$ on all datasets in the main paper.
We notice that $\sigma_{\mathrm{FFA}}=0.68$ works best on LPW while $\sigma_{\mathrm{FFA}}=0.71$ is the best on MARS.

%%%%%
\begin{table}[!h]
	\begin{center}
		\caption{
			The results of different $\sigma_{\mathrm{FFA}}$ on MARS and LPW in terms of CMC-1 (\%).
			We fix $\sigma_{\mathrm{FLA}}=0.18$ in all experiments. 
		}
		\label{table:different_sigma_FFA}
		\vspace{-8pt}
		\begin{tabular}{lcc}
			\hline
			$\sigma_{\mathrm{FLA}}=0.18$  & MARS & LPW \\
			\hline
			$\sigma_{\mathrm{FFA}}=0.62$ & 83.1 & 68.8 \\
			$\sigma_{\mathrm{FFA}}=0.65$ & 83.1 & 69.0 \\
			$\sigma_{\mathrm{FFA}}=0.68$ & 83.3 & 70.9 \\
			$\sigma_{\mathrm{FFA}}=0.71$ & 83.8 & 70.5 \\
			$\sigma_{\mathrm{FFA}}=0.74$ & 83.6 & 70.6 \\
			\hline
		\end{tabular}
	\end{center}
\end{table}
%%%%%

\end{document}